\documentclass[runningheads]{llncs}

\usepackage{tikz}
\usepackage{xcolor}
\usepackage{graphicx}

\begin{document}

\title{Exploiting In-Sensor Computing for Energy-Efficient Earth Observation}

\author{
Luigi Capogrosso\inst{1} \and
Pietro Bonazzi\inst{2} \and
Loris Hoxhaj\inst{3} \and
Michele Magno\inst{2,1}
}

\authorrunning{L. Capogrosso et al.}

\institute{
Interdisciplinary Transformation University of Austria, \email{name.surname@it-u.at}
\and
ETH Zurich, \email{name.surname@pbl.ee.ethz.ch}
\and
University of Verona, \email{name.surname@studenti.univr.it}}

\AddToHookNext{shipout/foreground}{%
\begin{tikzpicture}[overlay, remember picture]
    \node at ([yshift=-1.3cm]current page.north) {
        \normalsize\textcolor{gray}{This paper has been accepted for publication at the}
    };
    \node at ([yshift=-1.8cm]current page.north) {
        \normalsize\textcolor{gray}{XXIV Annual Conference on Sensors and Microsystems (AISEM), Catania, Italy, 2026}
    };
\end{tikzpicture}%
}

\maketitle

%%%%%%%%% ABSTRACT.
\begin{abstract}
The rapid growth of the satellite industry has driven a significant increase in geospatial data acquisition, highlighting a critical bottleneck: the severe disparity between the volume of collected sensor data and the limited downlink bandwidth available to ground stations.
While On-Board Computing (OBC) has helped address this by pre-processing data in orbit, this article further advances the paradigm by introducing an in-sensor computing framework.
We present an optimized end-to-end Earth Observation (EO) pipeline tailored for strict computational constraints by integrating TinyML techniques with the Sony IMX500 Intelligent Vision Sensor.
Specifically, our approach shifts processing directly to the sensor level, offloading the computation from the primary embedded device, and effectively mitigating the downlink transmission of noisy or irrelevant data.
We evaluated several efficient Convolutional Neural Networks (ConvNets), \emph{i.e.}, SqueezeNet, ShuffleNetV2, and MCUNetV1, on the EuroSAT dataset.
Experimental results show that, despite the optimizations required for deployment on the IMX500 platform, our models maintain a competitive 96.68\% accuracy while operating within its 8 MB constraints.
Specifically, the models reach an average processing throughput of 17.40 FPS with a latency of 27.43 ms.
Furthermore, our system profile exhibits high energy efficiency, with a low energy footprint of 14.19 mJ per inference and an efficiency rating of 42.26 GMAC/J, demonstrating its viability for in-sensor deployment.
\end{abstract}

%%%%%%%%% BODY TEXT.
\section{Introduction} \label{sec:intro}

During the past decade, the advent of New Space has democratized access to orbit \cite{Kodheli2021}.
Central to this shift are CubeSats, \emph{i.e.}, miniaturized satellites of a standardized $10 \times 10 \times 10$ cm form factor \cite{Liddle2020}, which now serve as the backbone for numerous Earth Observation (EO) missions.
These systems support a wide range of applications, including precision agriculture, maritime surveillance, and rapid disaster response \cite{Crisp2020}.
However, the inherent physical constraints of CubeSats in terms of size, weight, power consumption, and communication bandwidth fundamentally challenge the traditional operational model used for EO, which relies on downlinking most of the acquired sensor data for ground or cloud processing \cite{Eosdis2025}.

%%%%%%%%%%
\emph{\textbf{Motivations for this paper.}}
Specifically, the conventional EO workflow was designed for large-scale satellites equipped with continuous ground connectivity and high-capacity downlinks.
In this paradigm, raw Level-0 data is transmitted to ground stations and refined to higher-level products at centralized facilities \cite{Eosdis2025}.
Although highly effective for flagship missions, this transmit-everything approach is unsustainable for CubeSats.
Low-Earth orbit passes offer only narrow windows of contact \cite{Du2025}, and the energy cost per bit remains a critical bottleneck even with advanced S-band or X-band radios \cite{Babuscia2020}.

This bottleneck necessitates a paradigm shift: instead of moving all data to the ground, we must bring computation to the data.
This empowers satellites to autonomously filter, classify, and prioritize imagery, reducing downlink volume and latency for time-critical applications \cite{Capogrosso2026a}.
Achieving intelligence under the severe power, memory, and compute constraints of CubeSat-class platforms requires a strictly hardware-aware approach \cite{Capogrosso2024}. 
Although prior efforts have successfully migrated inference tasks to the satellite's main On-Board Computer (OBC) \cite{Capogrosso2026a}, or secondary accelerators \cite{Giuffrida2020,Giuffrida2022}, these solutions still tax the primary board's limited resources and necessitate power-hungry data transfers across the system bus. 

%%%%%%%%%%
\emph{\textbf{Scientific contribution.}}
As a result, this work introduces a paradigm shift: pushing computation to the absolute edge by executing deep learning models directly within the image sensor, significantly reducing the computational burden on the OBC, and the data downlink volume to ground stations.
In particular, we present an optimized end-to-end TinyML pipeline tailored specifically for in-orbit EO via in-sensor computing, using deployment-aware modifications such as static graph tracing and INT8 quantization.
Our optimized Convolutional Neural Networks (ConvNets), \emph{i.e.}, SqueezeNet \cite{Iandola2017}, ShuffleNetV2 \cite{Zhang2018}, and MCUNetV1 \cite{Lin2020}, achieve a competitive average accuracy of 96.68\% on the EuroSAT dataset \cite{Helber2019}.
These models run efficiently at 17 FPS with an average latency of 27.43 ms on the new Sony IMX500 intelligent vision sensor \cite{IMX500}, representing the first demonstration of real-time in-sensor data computation for EO missions.
Finally, these architectures are characterized by a high energy efficiency of 42.26 GMAC/J and a low per-inference consumption of 14.19 mJ, ensuring long-term operational sustainability.
\section{Related Works} \label{sec:related}

%%%%%%%%%%%%%%%%%%%%%%
\subsection{Efficient Deep Learning for Earth Observation}
The integration of deep learning into EO operations is rapidly becoming central to the operational design of modern missions.
Several surveys highlight this shift, emphasizing the necessity of onboard data processing to mitigate severe downlink bandwidth limitations and improve satellite autonomy \cite{Fourati2021,Miralles2021,Duggan2025,Chintalapati2025}.
A key driver for this integration is the pursuit of operational independence; for example, the FUTURE mission aims to advance the capabilities of autonomous spacecraft by leveraging Artificial Intelligence (AI) to reduce dependency on ground-station services \cite{Buonagura2023}.

Initial efforts focused on the implementation of highly targeted image-filtering systems to optimize communication resources \cite{Ghassemi2019}.
Similarly, the CloudScout project benchmarked an Intel Myriad 2 VPU on custom FPGA hardware, marking the first in-orbit use of ConvNets for hyperspectral cloud detection \cite{Giuffrida2020}.
These preliminary studies culminated in the European Space Agency (ESA) $\Phi$-Sat-1 mission, which achieved 96\% accuracy and a false-positive rate of less than 1\% on real space-acquired data \cite{Giuffrida2022}.
Building on this success, researchers of the $\Phi$-Sat-2 mission developed an AutoEncoder specifically optimized for the hardware onboard to perform lossy image compression, significantly reducing data volume while preserving essential image quality \cite{Guerrisi2023}.

To manage the massive influx of data from high-resolution, Low Earth Orbit sensors, researchers have proposed increasingly sophisticated edge computing frameworks.
In \cite{LeyvaMayorga2023}, the authors introduced a Satellite Mobile Edge Computing (SMEC) framework that optimizes image distribution and compression, increasing energy efficiency by a factor of 12.
From a hardware architecture point of view, in \cite{Zhang2022}, the authors developed a three-level system-on-chip-based edge computing design for the Luojia3 satellite, which accelerates data processing by more than two orders of magnitude compared to traditional methods.
Looking toward future constellations, the authors in \cite{Wang2025} recently proposed a distributed on-orbit cloud computing architecture.
This framework leverages ultra-heterogeneous system-on-chip platforms (combining CPUs, GPUs, FPGAs, and AI accelerators) to enable real-time collaborative processing across multiple satellites, significantly reducing ground reliance.

Despite these architectural advances, the deployment of deep learning models in space remains heavily constrained by harsh environmental factors, power limitations, and radiation.
In \cite{Castillo2024}, the authors highlighted these challenges and detailed design techniques to maximize the robustness of the system.
Consequently, recent literature extensively evaluates the trade-offs of various hardware platforms for Edge AI in space \cite{Diana2024,Mystkowska2025}.
In \cite{Capogrosso2026a}, the authors explored a hardware-aware TinyML pipeline for CubeSats, achieving task-acceptable accuracy while maintaining millijoule-level energy consumption per inference.
As noted in \cite{Duggan2025}, these strategies are essential for deploying lightweight models on high-performance hardware, ultimately enabling real-time applications such as the detection of natural disasters directly from orbit \cite{Chintalapati2025}.

%%%%%%%%%%%%%%%%%%%%%%
\subsection{Gaps in the Literature}
Despite progress in enabling onboard intelligence for EO, critical gaps remain unaddressed.
The majority of the existing work focuses on the deployment of learning models on primary OBC or dedicated accelerators \cite{Capogrosso2026a}, such as FPGAs or complex SoCs \cite{Giuffrida2020,Giuffrida2022}.
Although effective, this paradigm still requires the transfer of massive amounts of raw Level-0 data from the image sensor to the processing unit.
Consequently, this approach can introduce two equally critical bottlenecks: the massive internal data transfer saturates the satellite's bus and incurs a substantial dynamic power penalty; while subsequent inference tasks severely overload the OBC's limited processing unit.

In light of these challenges, the paradigm of in-sensor computing, where the vision sensor performs intelligent early-stage filtering before data are transmitted to the processing unit \cite{Capogrosso2026b}, represents a highly promising, yet unexplored frontier in the space sector.
This work bridges this gap by introducing the first fully optimized, architecture-based, in-sensor computing pipeline for EO missions. 
\section{Methodology} \label{sec:methodology}

%%%%%%%%%%%%%%%%%%%%%%
\subsection{Hardware Setup}
The target platform for our in-sensor computing framework is the Sony IMX500 intelligent vision sensor \cite{Eki2021,IMX500}.
The IMX500 is a 12.3-megapixel CMOS sensor (approximately $4056 \times 3040$ effective pixels, 1.55 $\mu$m pixel size) that uniquely integrates an Image Signal Processor (ISP), a dedicated Neural Network (NN) accelerator, and on-chip memory into a single stacked logic layer.
This architecture allows for the execution of quantized deep neural networks directly on the sensor die.
Consequently, instead of transmitting uncompressed raw images across the system bus, the sensor outputs only lightweight semantic metadata (\emph{e.g.}, classification labels or filtered region-of-interest coordinates).
This significantly reduces bandwidth utilization, system latency, and dynamic energy consumption in edge applications, as shown in \cite{Bonazzi2023,Bonazzi2025,Bonazzi2026a,Bonazzi2026b} and discussed in \cite{Capogrosso2026b}.

For our experimental validation, the IMX500 was evaluated using the Raspberry Pi AI Camera \cite{AICamera}, which interfaces with a Raspberry Pi 5 via the standard MIPI CSI-2 protocol.
In this configuration, the Raspberry Pi serves as an example of an OBC.
NN inference is performed entirely on the sensor, producing immediate image-level classifications, while the host OBC merely receives and aggregates the resulting metadata.
This setup perfectly simulates the architectural decoupling required for real-time and low-power data filtering in space environments.

%%%%%%%%%%%%%%%%%%%%%%
\subsection{Design Goals}
We restricted our architectural search to lightweight ConvNets for two primary reasons.
First, ConvNets provide an optimal trade-off between spatial feature extraction capabilities and parameter efficiency, which is critical for classifying complex EO imagery.
Second, the IMX500's embedded NN accelerator is explicitly designed and optimized for standard spatial operations, such as convolutions and element-wise activations, making ConvNets highly hardware-compliant compared to memory-intensive alternatives like Vision Transformers.

In particular, the optimization goals that guide our implementation and deployment are as follows:
\begin{enumerate}
    \item \textbf{Preserve performance:} Limit the accuracy drop resulting from quantization to a negligible and task-acceptable margin compared to the uncompressed Float32 baseline.
    \item \textbf{Strict memory compliance:} Fit all model parameters, intermediate layer activations, and working buffers within the highly constrained 8 MB embedded RAM of the IMX500.
    \item \textbf{Maximize throughput:} Achieve a high inference framerate to ensure that images can be processed in real-time.
    \item \textbf{Minimize data transfer:} Reduce communication overhead by processing raw Level-0 data directly on the sensor and transmitting only processed results to the OBC.
\end{enumerate}

%%%%%%%%%%%%%%%%%%%%%%
\subsection{Models Optimization}
To successfully deploy deep learning models onto the IMX500 sensor, the architecture must be processed through a fully hardware-compliant pipeline.
Therefore, we utilized the Model Compression Toolkit (MCT) \cite{SonyMCT}, an open-source framework released by Sony Semiconductor Solutions, specifically designed to optimize NNs on this target platform.
This toolkit provides advanced quantization, graph tracing, and compression utilities to bridge the gap between standard deep learning frameworks and embedded edge deployment.

Among these optimization techniques supported by the toolkit, we employed Post-Training Quantization (PTQ) to convert the network weights and activations from 32-bit floating-point (FP32) to 8-bit integer (INT8) representations.
PTQ was selected for its efficiency; it requires extremely low computational overhead and avoids the need for resource-intensive Quantization-Aware Training (QAT) loops \cite{Capogrosso2024}.
During the PTQ flow, the MCT's quantization core applies multiple algorithmic optimizations, such as operator fusion, optimal step-size calculation, and activation scaling, specifically tailored to the target IMX500 v1.0 hardware specifications \cite{SonyMCT}.
To calibrate the quantization parameters and minimize information loss, we supplied the MCT with a representative, unlabeled subset of the EuroSAT dataset.
This step ensures that the resulting INT8 models are statistically tuned to the statistical distributions characteristic of EuroSAT EO imagery.

Finally, we implemented an optimization step to address the architectural incompatibilities between the selected models and the IMX500 toolchain.
Specifically, we manually adjusted particular layer attributes in the post-quantization ONNX graphs to meet the platform's execution requirements, ensuring that all models remained compliant with the target hardware's inference engine.
\section{Experimental Results} \label{sec:experiments}

%%%%%%%%%%
\emph{\textbf{Datasets.}}
Our experimental evaluation is conducted on the EuroSAT dataset \cite{Helber2019}.
The primary reason for selecting this dataset is its role as a widely recognized state-of-the-art benchmark for EO and land-use classification, acquired during the Sentinel-2 ESA mission.
For this study, we utilized the RGB version, which comprises 27,000 labeled images with a resolution of $64 \times 64$ pixels, organized into 10 classes.

%%%%%%%%%%
\emph{\textbf{Models.}}
We used the following three models: SqueezeNet, ShuffleNetV2, and MCUNetV1.
The selection of these models was guided by two main criteria. 
First, rather than relying on memory-bound operations like self-attention, these networks utilize optimized convolutional building blocks, such as Fire modules in SqueezeNet, channel shuffling in ShuffleNetV2, and depthwise-separable convolutions in MCUNet.
These specific operations can be efficiently mapped to the sensor's INT8 arithmetic logic units and are fully supported by Sony's MCT. 
Second, despite their limited size, these models have proven capable of maintaining competitive feature extraction capabilities.
This ensures that they can effectively capture the complex spatial patterns necessary for accurate EO tasks, even after different model optimizations.

%%%%%%%%%%
\emph{\textbf{Training details.}}
The models were implemented using PyTorch \cite{Paszke2019}.
We trained our models for 50 epochs, using Adam as optimizer \cite{Kingma2015}, a learning rate of $1 \times 10^{-3}$, and a batch size of 32.
We used an 80/20 train-test split for the training and test sets, respectively.

%%%%%%%%%%
\emph{\textbf{Evaluation setup.}}
We evaluated the performance of the models deployed on the IMX500 across four key evaluation metrics: classification accuracy, inference latency, memory occupancy, and energy consumption.

%%%%%%%%%%%%%%%%%%%%%%
\subsection{Accuracy, Memory, and Energy Analysis}
\begin{figure}[t!]
    \centering
    \includegraphics[width=\linewidth]{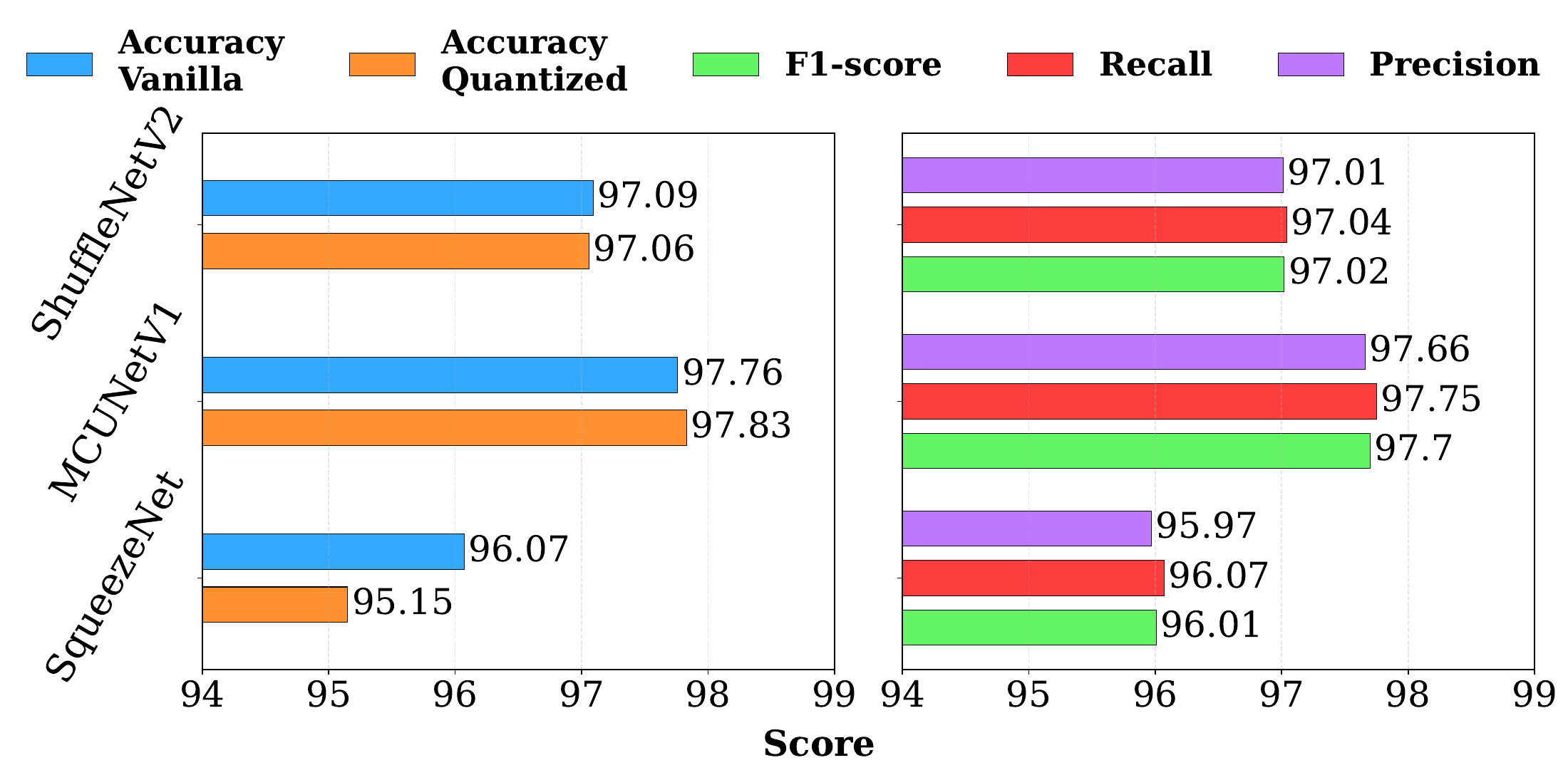}
    \caption{Performance comparison of ShuffleNetV2, MCUNetV1, and SqueezeNet.
    The left image illustrates the comparisons in terms of classification accuracy for both FP32 and INT8 models.
    The right panel displays the F1-score, Recall, and Precision metrics for the vanilla versions.}
    \label{fig:results}
\end{figure}

The performance of the selected models was evaluated by focusing on classification accuracy, reliability metrics, impact of quantization, and hardware-level efficiency.

%%%%%%%%%%
\emph{\textbf{Classification accuracy.}}
MCUNetV1 emerged as the most efficient model, achieving an accuracy of 97.76\%.
ShuffleNetV2 followed closely with 97.09\%.
Finally, SqueezeNet scored the lowest overall among the candidates, although it remained a competitive baseline with a original precision of 96.07\%.

%%%%%%%%%%
\emph{\textbf{Reliability metrics.}}
The F1-score, Recall, and Precision confirmed a balanced classification performance across all models, ensuring that there was no significant class-specific bias.
MCUNetV1 maintained the most balanced performance with an F1-score of 97.7\%.
ShuffleNetV2 demonstrated near-perfect symmetry, with precision and recall staying near 97.0\%, while SqueezeNet remained robust with an F1-score of 96.01\%, confirming its strong generalization capabilities despite its lower overall accuracy.

%%%%%%%%%%
\emph{\textbf{Impact of quantization.}}
MCUNetV1 and ShuffleNetV2 demonstrated remarkable resilience to bit-width reduction.
In particular, ShuffleNetV2 showed a negligible drop of only 0.03\%, while MCUNetV1 actually exhibited a marginal improvement, reaching 97.83\%, suggesting that the process acted as a form of weight regularization in this case.
SqueezeNet proved to be the most susceptible to PTQ, with a performance degradation of 0.92\%, which reduced its accuracy to 95.15\%.

%%%%%%%%%%
\emph{\textbf{Hardware-level efficiency.}}
We evaluated the memory footprint in MB and the processing throughput in FPS to quantify real-time performance.
Furthermore, to assess energy sustainability, we measured the millijoules (mJ) consumed per inference and the Giga-Multiply-Accumulate operations per Joule (GMAC/J).
On average, our models achieved a processing throughput of 17.40 FPS, maintaining an average latency of 27.43 ms across 100 processed frames.
Regarding energy efficiency, the models consumed an average of 14.19 mJ per inference (at an assumed power of 0.247 W).
This translates to a computational efficiency of 10.443 GMAC/s and a sustainability metric of 42.26 GMAC/J.
\section{Conclusions \& Future Work} \label{sec:conclusions}

In this article, we propose a novel perspective to address the critical communication and computational bottlenecks that constrain modern CubeSat-based platforms.
By shifting the deep learning workload away from the primary OBC and directly into the image sensor, we introduced the first optimized in-sensor computing pipeline for EO missions.
This paradigm shift effectively resolves two major challenges: it drastically reduces the volume of irrelevant data transmitted over the limited downlink bandwidth, and it eliminates the dynamic power penalty and latency associated with transferring raw Level-0 imagery across the satellite's internal data bus.

To validate this approach, we used the Sony IMX500 intelligent vision sensor and its native MCT.
Our end-to-end TinyML pipeline successfully applied hardware-aware PTQ to convert standard FP32 architectures into efficient INT8 models.
Experimental results on the EuroSAT dataset demonstrate that the optimized ConvNets maintain high accuracy, achieving an average accuracy of 96.68\% while operating strictly within the embedded 8 MB RAM constraints of the IMX500.
Our models achieve a throughput of 17.40 FPS with a latency of 27.43 ms.
By consuming only 14.19 mJ per inference at an efficiency of 42.26 GMAC/J, they ensure long-term operational sustainability.

Future work will focus on validating the radiation tolerance of these integrated sensing platforms, as well as exploring on-orbit learning techniques to dynamically update the quantized models without requiring full retraining from the ground.

\section*{Acknowledgments}
The research was funded by the Swiss National Science Foundation (Grant 219943).

%%%%%%%%% BIBLIOGRAPHY.
\bibliographystyle{plain}
\bibliography{bibliography}

\end{document}